\documentclass{article} %
\usepackage{arxiv,times}

\usepackage{amsmath,amsfonts,bm}

\def\eqref#1{equation~\ref{#1}}

\def\1{\bm{1}}

\DeclareMathAlphabet{\mathsfit}{\encodingdefault}{\sfdefault}{m}{sl}
\SetMathAlphabet{\mathsfit}{bold}{\encodingdefault}{\sfdefault}{bx}{n}

\usepackage{hyperref}
\usepackage{url}
\usepackage{booktabs}
\usepackage[capitalise]{cleveref} 
\usepackage{graphicx}
\usepackage{wrapfig}
\usepackage{capt-of}

\title{AffordDrive3D: Affordance-Aware World-Action Modeling with Spatial Understanding}

\makeatletter
\renewcommand{\headrulewidth}{0pt}
\def\@maketitle{\vbox{\hsize\textwidth\centering
  {\LARGE\bf \@title\par}
  \vskip 0.25in
  \@author\par}
  \vskip 0.3in minus 0.1in}
\makeatother

\author{
\textbf{Tianhui Cai}$^{1,2*\ddagger}$\quad
\textbf{Xinglong Sun}$^{2*}$\quad
\textbf{Chao Fang}$^{3}$\quad
\textbf{Zhenxin Li}$^{2,4\ddagger}$\quad
\textbf{Rui Song}$^{1}$\\[2pt]
\textbf{Jose M. Alvarez}$^{2}$\quad
\textbf{Yunxiang Mao}$^{2}$\quad
\textbf{Jiaqi Ma}$^{1}$\quad
\textbf{Langechuan Liu}$^{2\dagger}$\\[6pt]
{\normalfont\small
$^1$University of California, Los Angeles \quad $^2$NVIDIA \quad $^3$42dot by Hyundai \quad $^4$University of Toronto}
}

\arxivfinalcopy
\begin{document}

\maketitle
{\renewcommand{\thefootnote}{}\footnotetext{$^{*}$Equal contribution.\quad $^{\dagger}$Corresponding author.\quad $^{\ddagger}$Work done during an internship at NVIDIA.}}
\begin{abstract}
World-action models have recently improved autonomous driving by jointly learning future scene prediction and trajectory generation. Most existing approaches model the future primarily through RGB appearance, and recent works have begun to incorporate geometric prediction to improve spatial understanding. However, dense geometry describes the spatial layout of the entire scene without indicating which parts are most relevant to the ego vehicle's action. For driving, the model must also identify and anticipate where it can safely move and which regions may pose collision risks. Jointly modeling action-relevant regions and future geometry can provide the policy with both driving-relevant cues and their corresponding spatial structure. We therefore propose \textbf{AffordDrive3D}, an affordance- and geometry-aware world-action model that jointly learns future action-relevant regions and spatial structure. In order to capture the scene semantics and driving context needed for driving affordance prediction, we build AffordDrive3D on a VLM backbone to forecast drivable areas and collision-critical regions that directly affect ego motion, while predicting future geometry from RGB world-model latents. On NAVSIM, AffordDrive3D achieves state-of-the-art performance with 91.3 PDMS and 89.9 EPDMS, demonstrating the effectiveness of jointly modeling future affordances and geometry for trajectory planning. 

\textbf{Project Page:} \url{https://afforddrive3d.github.io/}

\end{abstract}

\section{Introduction}

Autonomous driving has increasingly shifted toward end-to-end models that directly map sensor observations to future trajectories, allowing scene representations and driving policies to be jointly optimized for motion planning \citep{hu2023planning, jiang2023vad,liao2025diffusiondrive,li2025generalized,li2025ztrs,yao2026had,li2026large}. More recently, advances in vision-language models (VLMs) have demonstrated strong semantic understanding and reasoning capabilities for complex driving scenes. Vision-language-action (VLA) models further incorporate action prediction into this framework, enabling a unified model to interpret the driving scene, reason about its context, and generate future ego actions \citep{zhou2025autovla,zhou2026opendrivevla,renz2025simlingo, fu2025orion}.

However, driving decisions depend not only on understanding the current observation, but also on anticipating how the environment may evolve. Motivated by this need, world-action models (WAMs) further couple action learning with future-scene prediction, allowing a shared representation to jointly model how the environment evolves and how the ego vehicle should act \citep{zhang2026resworld, li2025end, hong2026drivefuture}. One line of work extends VLA models with explicit world modeling, introducing future prediction as additional supervision for action learning \citep{li2026drivevla, wang2026learning, liu2026driveworld}. Another line builds on pretrained video foundation models, adapting either learned temporal representations or generative video priors for joint future prediction and action learning \citep{shi2026drivewam, zhao2026simwam, wang2026wa}. By supervising the model with future observations in addition to expert trajectories, WAMs encourage the policy to capture scene dynamics that are relevant to future decision making.

Most existing WAMs model the future primarily through RGB appearance or visual features \citep{wang2024driving, zhao2026forecasting, chen2025drivinggpt, li2025enhancing}. While such prediction provides rich supervision for scene evolution, appearance alone does not explicitly represent the spatial structure needed for driving. Compared with RGB appearance, geometry provides a more direct representation of scene structure, explicitly capturing relative distance and spatial layout that must otherwise be inferred from visual appearance. Recent approaches have therefore begun to incorporate geometric prediction into future modeling, such as depth \citep{sheng2026explorevla,zhang2026geoworldad}, point cloud \citep{lu2026geowam} and 3D occupancy \citep{yang2025driving, cheng2026owmdrive}. However, geometric future modeling remains relatively underexplored in VLA-based driving frameworks \citep{sheng2026explorevla}.

Geometric future modeling provides explicit spatial structure that is missing from appearance-based prediction, yet a key challenge remains: how to use future spatial information to improve policy learning and action prediction. In autonomous driving, this requires identifying which regions of the future scene are relevant to ego motion, such as drivable areas and regions that may pose collision risks. Without capturing their relevance to action, future world state prediction remains insufficiently aligned with the planning objective.

\begin{figure}[t]
    \centering
    \includegraphics[width=0.95\linewidth]{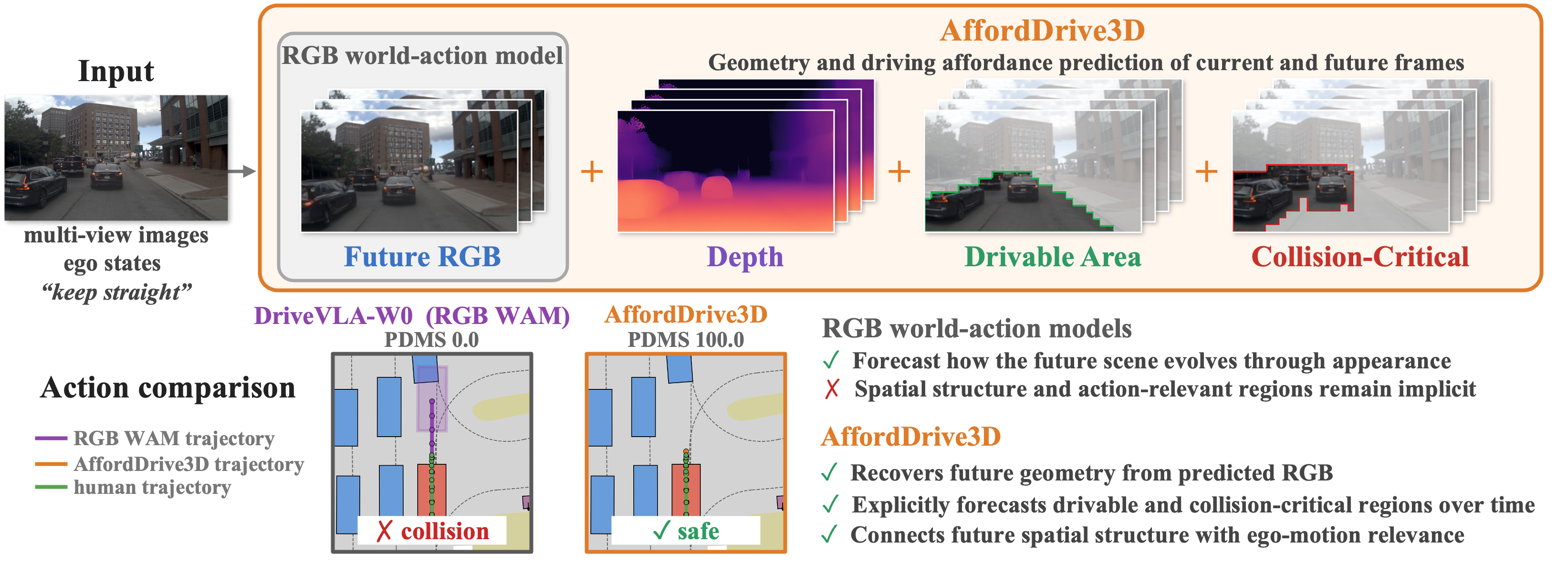}
        \caption{\textbf{AffordDrive3D vs.\ RGB-based world-action models.} Existing RGB-based world-action models primarily capture future scene evolution through appearance prediction. AffordDrive3D additionally predicts future geometry to model how spatial structure evolves, alongside driving affordances that identify spatial regions critical to the ego vehicle’s motion. Together, these predictions align future world state modeling more closely with planning objectives and improve trajectory prediction.}
        
    \label{fig:teaser}
    \vspace{-0.4cm}
\end{figure}

Therefore, we propose \textbf{AffordDrive3D}, an affordance- and geometry-aware world-action model that jointly learns future action-relevant regions and spatial structure as shown in \cref{fig:teaser}. For geometric modeling, we condition depth prediction on the future RGB latents produced by the world model. Because these latents already encode how the scene changes over time, the geometry branch does not need to independently predict future dynamics and can instead focus on recovering the corresponding 3D structure, while keeping the RGB and geometric predictions grounded in the same future scene. Beyond spatial structure, we further model which regions of the scene are important for driving. We refer to these action-relevant regions as \textbf{\textit{driving affordances}}, including drivable areas and collision-critical regions, and introduce temporal affordance forecasting to predict them at both current and future frames. To provide the contextual scene understanding needed for affordance prediction, we adopt a VLM backbone and leverage its strong semantic understanding and reasoning capabilities. We jointly learn affordance prediction, future RGB and depth modeling, and trajectory generation through a shared representation. Compared with ExploreVLA \citep{sheng2026explorevla}, which also incorporates future depth prediction, AffordDrive3D improves PDMS by 0.9 points and EPDMS by 1.1 points on the NAVSIM benchmark \citep{dauner2024navsim}, achieving new state-of-the-art (SOTA) scores of 91.3 PDMS and 89.9 EPDMS.

Our main contributions are:
\begin{itemize}

\item We propose \textbf{AffordDrive3D}, an affordance- and geometry-aware WAM built on a VLA framework, which extends future world modeling beyond visual appearance to better align it with the planning objective.
\item We jointly model future geometry and temporal driving affordances: RGB-conditioned future depth captures the spatial structure of the predicted scene, while affordance forecasting identifies drivable areas and collision-critical regions, connecting future spatial information to ego motion.
\item We achieve new SOTA performance with 91.3 PDMS and 89.9 EPDMS on the NAVSIM benchmark. Ablation studies further demonstrate the effectiveness of temporal affordance forecasting and future geometry modeling for trajectory planning.
\end{itemize}

\section{Related Work}
\subsection{VLA and WAM for Autonomous Driving}

VLMs have increasingly been adopted for autonomous driving to integrate perception, reasoning, and decision-making. DriveLM~\citep{sima2024drivelm} introduces graph-structured visual question answering, while DriveVLM~\citep{tian2024drivevlm} integrates multimodal reasoning with planning. Recent VLA frameworks, including AutoVLA~\citep{zhou2025autovla}, HybridDriveVLA~\citep{bassole2026hybriddrivevla}, and ColaVLA~\citep{peng2026colavla}, further connect multimodal reasoning with trajectory generation through action tokenization or visual or latent reasoning. However, reasoning over observed scenes alone does not explicitly model how the environment evolves over time.

World models address this limitation by predicting future observations or structured scene states. Drive-WM~\citep{wang2024driving} explores action-conditioned video generation, while Drive-OccWorld~\citep{yang2025driving} predicts future occupancy and flow. More recently, WAMs jointly model future dynamics and actions: DriveWAM~\citep{shi2026drivewam} adapts video generative priors to joint video-action generation, WA-JEPA~\citep{wang2026wa} learns latent dynamics with ego trajectories, and SimWAM~\citep{zhao2026simwam} as well as SV-WAM~\citep{wang2026sv} use video prediction as training-time supervision. These approaches improve predictive action learning, but often rely on RGB, latent, or generic geometric dynamics without explicitly structuring future states around action-relevant spatial constraints.

\subsection{Geometry-Aware Driving Policies }

Explicit geometry provides spatial information complementary to images. Drive-OccWorld~\citep{yang2025driving} models future occupancy and flow, GenieDrive~\citep{yang2026geniedrive} uses 4D occupancy to guide video generation, and SpaceDrive~\citep{li2026spacedrive} incorporates depth-derived 3D information into VLM-based driving. ExploreVLA~\citep{sheng2026explorevla} predicts future RGB and depth in parallel for trajectory planning.

Recent work also explores geometric dynamics directly within world-action modeling. In particular, GeoWAM~\citep{lu2026geowam} directly forecasts future point-cloud geometry for action learning. In contrast, our model predicts future depth conditioned on RGB latents from the diffusion world model and jointly learns RGB, depth, affordance, and trajectory objectives, coupling anticipated appearance with geometric structure. Still, geometry alone does not explicitly indicate which regions are drivable or collision-critical for ego motion.

\subsection{Affordance-Aware Planning}

Affordances provide a task-relevant abstraction between perception and action by emphasizing regions and interactions that directly influence control. In robotics, VLAs use task-conditioned or structured affordances to bridge perception and action~\citep{wang2026afford,yu2026affordancevla}. Affordance-aware learning remains less explored in autonomous driving. Recent methods incorporate structured future priors or drivable-area constraints for planning~\citep{zeng2026futuresightdrive,wang2026sv}. Rather than relying on future visual reasoning, structured scene priors, or trajectory-level constraints, we explicitly forecast current and future drivable and collision-critical regions, which provides temporally structured supervision for how action-relevant space evolves over time.

\section{Method}

\begin{figure}[t]
    \centering
    \includegraphics[width=0.95\linewidth]{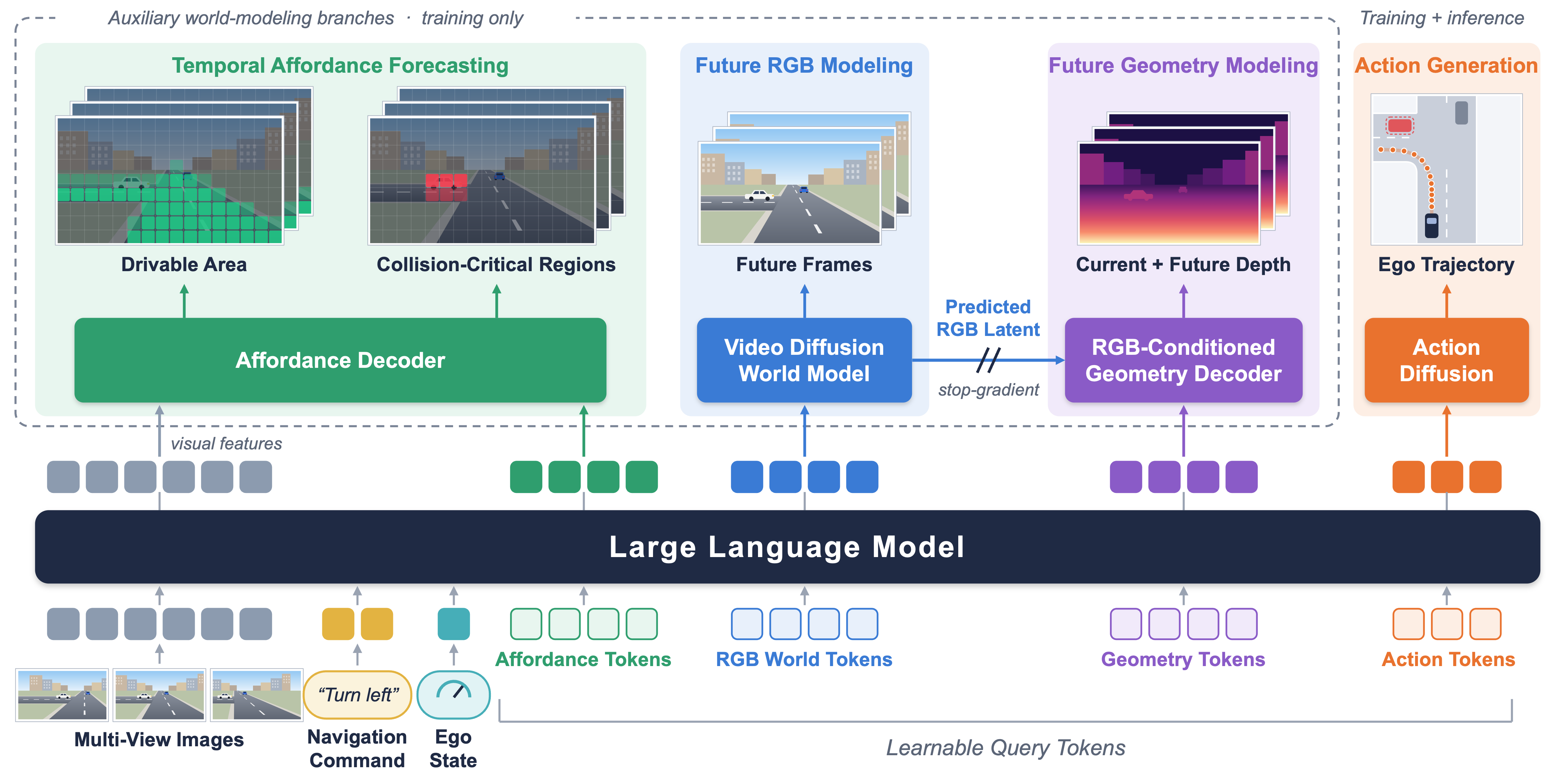}
    \caption{\textbf{Framework of AffordDrive3D.}
Multi-view images, the navigation command, and the ego state are encoded by a shared VLM backbone together with learnable affordance, RGB world, geometry, and action tokens. The auxiliary branches (dashed box) forecast driving affordances, future RGB, and RGB-conditioned depth to shape the shared representation during training, while only the action branch is used at inference.}
    \label{fig:framwork}
    \vspace{-0.4cm}
\end{figure}
\subsection{Overview}

As shown in \cref{fig:framwork}, given the current multi-view observations $\mathcal{I}_t=\{I_t^{c}\}_{c=1}^{C}$, a navigation command $g$, and the ego state $s$, AffordDrive3D predicts a future ego trajectory $\hat{\tau}_{t:t+K}$ while jointly modeling driving affordances, scene appearance, and geometry across the current and future scene. 
We use a shared causal VLM backbone to encode the inputs together with four groups of learnable tokens: affordance tokens $\mathbf{Q}^{\mathrm{aff}}$, RGB world tokens $\mathbf{Q}^{\mathrm{rgb}}$, geometry tokens $\mathbf{Q}^{\mathrm{geo}}$, and action tokens $\mathbf{Q}^{\mathrm{act}}$. $\mathbf{Q}^{\mathrm{aff}}$ model current and future drivable and collision-critical regions, $\mathbf{Q}^{\mathrm{rgb}}$ condition the diffusion world model to forecast future visual appearance, and $\mathbf{Q}^{\mathrm{geo}}$ predict current and future scene geometry with guidance from the predicted RGB latent. Finally, the action tokens $\mathbf{Q}^{\mathrm{act}}$ condition an action diffusion model to generate the future ego trajectory. 

All components are optimized jointly through the shared backbone, allowing affordance forecasting, visual prediction, and geometry modeling to shape the representation used for planning. During inference, only the action branch is required to generate the ego trajectory. The auxiliary affordance, RGB, and depth outputs do not need to be decoded.

\subsection{RGB-Conditioned Future Geometry Modeling}

\paragraph{Future RGB modeling.}
The RGB world tokens $\mathbf{Q}^{\mathrm{rgb}}$ produce hidden states $\mathbf{H}^{\mathrm{rgb}}$ after the shared causal backbone, which condition a video diffusion model to predict future scene appearance in latent space. We encode the current and future RGB frames using a video VAE, obtaining a sequence of  $T$ temporal latent groups $\mathbf{Z}^{\mathrm{rgb}}=[\mathbf{Z}^{\mathrm{rgb}}_{0},\ldots,\mathbf{Z}^{\mathrm{rgb}}_{T-1}]$ after temporal compression, where $\mathbf{Z}^{\mathrm{rgb}}_{0}$ corresponds to the current-frame anchor and $\mathbf{Z}^{\mathrm{rgb}}_{1:T-1}$ represents the future sequence.

Following flow-matching, we perturb the full video latent, including the anchor group, as
\begin{equation}
\mathbf{Z}^{\mathrm{rgb}}_{\sigma}
=
(1-\sigma)\mathbf{Z}^{\mathrm{rgb}}
+
\sigma\boldsymbol{\epsilon},
\end{equation}
where $\boldsymbol{\epsilon}$ is sampled noise and $\sigma$ denotes the noise level. The video diffusion model is set up as an inpainting model, where the current visual observation is additionally encoded as a conditioning signal $\mathbf{C}_{0}$. The model predicts the velocity field as
\begin{equation}
\mathbf{v}_{\theta}
=
\mathcal{W}_{\mathrm{rgb}}
\left(
\mathbf{Z}^{\mathrm{rgb}}_{\sigma},
\sigma;
\mathbf{H}^{\mathrm{rgb}},
\mathbf{C}_{0}
\right),
\end{equation}
where $\mathbf{H}^{\mathrm{rgb}}$ provides the VLM context for future generation, while $\mathbf{C}_{0}$ provides the observed current-frame appearance. The RGB world model is trained with the flow-matching objective
\begin{equation}
\mathcal{L}_{\mathrm{rgb}}
=
\left\|
\mathbf{v}_{\theta}
-
\left(
\boldsymbol{\epsilon}
-
\mathbf{Z}^{\mathrm{rgb}}
\right)
\right\|_2^2.
\end{equation}
From the predicted velocity, we recover the predicted clean RGB latent as
\begin{equation}
\hat{\mathbf{Z}}^{\mathrm{rgb}}
=
\mathbf{Z}^{\mathrm{rgb}}_{\sigma}
-
\sigma\mathbf{v}_{\theta},
\end{equation}
which is subsequently used as future-scene context for geometry prediction.

\paragraph{RGB-conditioned geometry prediction.}
The RGB world model already predicts how the visual scene evolves over future frames. We use the denoised RGB latent $\hat{\mathbf{Z}}^{\mathrm{rgb}}$ as future-scene context for geometry prediction \citep{liang2025unifuture}, allowing the geometry branch to focus on recovering the 3D structure associated with the predicted visual evolution.

We first map the geometry-token hidden states $\mathbf{H}^{\mathrm{geo}}$ to a coarse temporal-spatial feature grid $\mathbf{G}^{\mathrm{geo}}\in\mathbb{R}^{T_g\times H_g\times W_g\times D}$, where $T_g$, $H_g$, and $W_g$ denote the temporal, height, and width dimensions of the grid, respectively, with $T_gH_gW_g=|\mathbf{Q}^{\mathrm{geo}}|$. We then use trilinear interpolation to upsample the coarse geometry feature grid $\mathbf{G}^{\mathrm{geo}}$ to the same temporal-spatial resolution $(T,H,W)$ as the predicted RGB latent $\hat{\mathbf{Z}}^{\mathrm{rgb}}$, yielding $\tilde{\mathbf{G}}^{\mathrm{geo}}\in\mathbb{R}^{T\times H\times W\times D}$. 

To condition geometry prediction on the predicted future scene, we apply a 3D convolution $\phi_{\mathrm{rgb}}$ to $\hat{\mathbf{Z}}^{\mathrm{rgb}}$ to map its feature channels to dimension $D$ while preserving its temporal-spatial resolution. The resulting RGB features are fused with $\tilde{\mathbf{G}}^{\mathrm{geo}}$ at corresponding temporal-spatial locations:
\begin{equation}
\mathbf{F}^{\mathrm{geo}}
=
\mathcal{R}_{\mathrm{3D}}
\left(
\phi_{\mathrm{fuse}}
\left(
\left[
\tilde{\mathbf{G}}^{\mathrm{geo}};
\phi_{\mathrm{rgb}}
\left(
\operatorname{sg}\!\left(\hat{\mathbf{Z}}^{\mathrm{rgb}}\right)
\right)
\right]
\right)
\right),
\end{equation}
where $\phi_{\mathrm{fuse}}$ denotes a $1\times1\times1$ 3D convolution that fuses RGB and geometry features, and $\mathcal{R}_{\mathrm{3D}}$ denotes a stack of 3D residual blocks for further feature refinement. The stop-gradient operator $\operatorname{sg}(\cdot)$ prevents the geometry loss from back-propagating through the RGB prediction branch.

The fused representation $\mathbf{F}^{\mathrm{geo}}$ retains the anchor--future temporal organization of the RGB latent space. Its anchor component $\mathbf{F}^{\mathrm{geo}}_{0}$ is formed with RGB context grounded in the observed current frame, whereas the future components $\mathbf{F}^{\mathrm{geo}}_{1:T-1}$ are formed with RGB context predicted by the RGB diffusion model. This yields two distinct geometry prediction settings: recovering current geometry from observation-grounded RGB context and predicting future geometry from predicted RGB context. We therefore use separate decoders for the anchor and future depth latents:
\begin{equation}
\hat{\mathbf{Z}}^{\mathrm{dep}}_{0}
=
\mathcal{D}_{\mathrm{anc}}
\left(
\mathbf{F}^{\mathrm{geo}}_{0}
\right),
\qquad
\hat{\mathbf{Z}}^{\mathrm{dep}}_{1:T-1}
=
\mathcal{D}_{\mathrm{fut}}
\left(
\mathbf{F}^{\mathrm{geo}}_{1:T-1}
\right).
\end{equation}

We encode the pseudo-ground-truth depth from Depth Anything 3 \citep{depthanything3} with the video VAE to obtain the target depth latent $\mathbf{Z}^{\mathrm{dep}}$, and supervise the predicted depth latent $\hat{\mathbf{Z}}^{\mathrm{dep}}$ with an $\mathcal{L}_1$ loss:
\begin{equation}
\mathcal{L}_{\mathrm{geo}}
=
\lambda_{\mathrm{anc}}
\mathcal{L}_{1}
\left(
\hat{\mathbf{Z}}^{\mathrm{dep}}_{0},
\mathbf{Z}^{\mathrm{dep}}_{0}
\right)
+
\mathcal{L}_{1}
\left(
\hat{\mathbf{Z}}^{\mathrm{dep}}_{1:T-1},
\mathbf{Z}^{\mathrm{dep}}_{1:T-1}
\right),
\end{equation}
where $\lambda_{\mathrm{anc}}$ controls the relative weight of the anchor-depth loss.

\subsection{Temporal Affordance Forecasting}

We index the affordance tokens by prediction horizon and camera view, such that $\mathbf{Q}^{\mathrm{aff}}=\{\mathbf{q}^{\mathrm{aff}}_{k,c}\}$, where $k\in\{0,\ldots,K\}$ denotes the future timestamp for prediction and $c\in\{1,\ldots,C\}$ denotes the camera view. After processing by the shared causal backbone, these tokens produce the corresponding hidden states $\mathbf{H}^{\mathrm{aff}}=\{\mathbf{h}^{\mathrm{aff}}_{k,c}\}$.

\paragraph{Dense affordance prediction.} Each affordance state $\mathbf{h}^{\mathrm{aff}}_{k,c}$ is decoded into two spatial predictions: a drivable-area map $\hat{\mathbf{Y}}^{\mathrm{da}}_{k,c}$ and a collision-critical map $\hat{\mathbf{Y}}^{\mathrm{col}}_{k,c}$. Both maps are defined over the visual-token grid of camera $c$, so each map value corresponds to one local patch of the image.

Let $\mathbf{v}_{c,n}$ denote the hidden state of the $n$-th visual token from camera $c$. For each affordance type $m\in\{\mathrm{da},\mathrm{col}\}$, we compute

\begin{equation}
z^{m}_{k,c,n}=
\frac{\left\langle f^{m}(\mathbf{h}^{\mathrm{aff}}_{k,c}),g^{m}(\mathbf{v}_{c,n})\right\rangle}{\sqrt{d}}+\mathrm{bias}^{m},
\end{equation}

where $f^{m}$ and $g^{m}$ are learnable projections that map the affordance and visual features into a $d$-dimensional embedding space. The predicted affordance probability at each visual-token location is then given by $\hat{Y}^{m}_{k,c,n}=\mathrm{sigmoid}(z^{m}_{k,c,n})$. The visual features provide current-scene context, while the frame-specific affordance states determine how the affordance predictions evolve over time.

\paragraph{Affordance supervision.} We construct drivable-area and collision-critical targets that are directly relevant to ego planning, focusing on reachable road regions and objects that may collide with the ego vehicle. The drivable-area target $Y^{\mathrm{da}}_{k,c,n}\in[0,1]$ measures the fraction of verifiable ground within a visual-token region that belongs to the road surface reachable by the ego vehicle, with intermediate values capturing road boundaries. To construct the collision-critical target $Y^{\mathrm{col}}_{k,c,n}\in\{0,1\}$, we select plausible ego trajectories from a predefined trajectory set \citep{li2024hydra} and roll them out in the scene. If any selected trajectory results in a collision with an agent or object, the visual-token region containing the collided agent or object is marked as collision-critical. We provide details of the label construction in Appendix~\ref{app:affordance}.

For drivable-area prediction, we use weighted soft binary cross-entropy,
\begin{equation}
\mathcal{L}^{k}_{\mathrm{da}}
=
\mathrm{WBCE}\left(
\mathbf{z}^{\mathrm{da}}_{k},
\mathbf{Y}^{\mathrm{da}}_{k};
\mathbf{w}^{\mathrm{da}}_{k}
\right),
\end{equation}
where $\mathbf{w}^{\mathrm{da}}_{k}$ down-weights tokens with limited ground evidence, such as those dominated by sky or buildings, and gives higher weight to tokens that contain drivable-area boundaries. In this way, the loss focuses more on reliable ground regions while preserving the fine-grained boundary information encoded by the soft targets.

For collision-critical prediction, we combine class-normalized BCE to address the class imbalance between positive and negative tokens with Dice loss to encourage spatial overlap between the predicted and ground-truth critical regions,

\begin{equation}
\mathcal{L}^{k}_{\mathrm{col}}
=
\mathrm{BCE}\left(
\mathbf{z}^{\mathrm{col}}_{k},
\mathbf{Y}^{\mathrm{col}}_{k}
\right)
+
\lambda_{\mathrm{dice}}
\mathrm{Dice}\left(
\hat{\mathbf{Y}}^{\mathrm{col}}_{k},
\mathbf{Y}^{\mathrm{col}}_{k}
\right).
\end{equation}

Similar to geometry prediction, we separate the affordance of the observed current frame from that of the forecast future frames and aggregate the losses as
\begin{equation}
\mathcal{L}_{\mathrm{aff}}=
\lambda_{\mathrm{anc}}\left(\mathcal{L}^{0}_{\mathrm{da}}+\mathcal{L}^{0}_{\mathrm{col}}\right)
+\frac{1}{K}\sum_{k=1}^{K}\left(\mathcal{L}^{k}_{\mathrm{da}}+\mathcal{L}^{k}_{\mathrm{col}}\right).
\end{equation}

\subsection{Action Generation and Joint Learning}

We use the hidden states of the action tokens $\mathbf{H}^{\mathrm{act}}$ to condition a DiT for future trajectory generation. The DiT is trained with a flow-matching objective $\mathcal{L}_{\mathrm{act}}$ to predict the velocity field between Gaussian noise and the ground-truth trajectory.

To encourage the RGB and geometry tokens to preserve action-relevant information while modeling future appearance and spatial structure, we additionally apply trajectory supervision to both the RGB and geometry tokens. Specifically, $\mathbf{H}^{\mathrm{rgb}}$ and $\mathbf{H}^{\mathrm{geo}}$ are separately processed by lightweight GRU-based trajectory heads to predict the future ego trajectory. We supervise these predictions with the ground-truth trajectory using an $\mathcal{L}_1$ loss, denoted as $\mathcal{L}_{\mathrm{rgb\text{-}act}}$ and $\mathcal{L}_{\mathrm{geo\text{-}act}}$, respectively. 

The complete framework is trained end-to-end with
\begin{equation}
\mathcal{L}=
\lambda_{\mathrm{aff}}\mathcal{L}_{\mathrm{aff}}
+\lambda_{\mathrm{rgb}}\mathcal{L}_{\mathrm{rgb}}
+\lambda_{\mathrm{geo}}\mathcal{L}_{\mathrm{geo}}
+\lambda_{\mathrm{act}}\mathcal{L}_{\mathrm{act}}
+\lambda_{\mathrm{rgb\text{-}act}}\mathcal{L}_{\mathrm{rgb\text{-}act}}
+\lambda_{\mathrm{geo\text{-}act}}\mathcal{L}_{\mathrm{geo\text{-}act}},
\label{eq:joint_loss}
\end{equation}
where $\lambda$ denotes the loss weights.

\section{Experiments}
\subsection{Experimental Setup}
\paragraph{Dataset and Metrics.}
We train and evaluate AffordDrive3D on the NAVSIM benchmark \citep{dauner2024navsim}, which is built on OpenScene and provides approximately 103K training samples in \texttt{navtrain} and 12K evaluation samples in \texttt{navtest}. We report both the PDM Score (PDMS) from NAVSIM v1 and the Extended PDM Score (EPDMS) from NAVSIM v2. NAVSIM v1 PDMS combines No At-Fault Collisions (NC), Drivable Area Compliance (DAC), Ego Progress (EP), Time-to-Collision (TTC), and Comfort (Comf.). NAVSIM v2 EPDMS further includes Driving Direction Compliance (DDC), Traffic Light Compliance (TLC), Lane Keeping (LK), History Comfort (HC), and Extended Comfort (EC). We follow the official NAVSIM evaluation protocol for both metrics.

For temporal affordance prediction, we additionally report average precision (AP). For drivable-area prediction, regions with a ground-truth drivable ratio greater than $0.5$ are treated as drivable. Further details on the AP computation are provided in Appendix~\ref{app:ap}.

\paragraph{Training and Implementation Details.}

We use Qwen3-VL-2B \citep{bai2025qwen3} as the VLM backbone and instantiate the RGB diffusion world model with Wan2.1-Fun-1.3B-InP \citep{wan2025}. We use $C=3$ camera views and set $K=8$ at 2\,Hz, corresponding to a 4\,s prediction horizon and $T=3$ latent groups after VAE encoding. We use $|\mathbf{Q}^{\mathrm{aff}}|=C(K+1)=27$ affordance tokens, $|\mathbf{Q}^{\mathrm{rgb}}|=64$ RGB world tokens, $|\mathbf{Q}^{\mathrm{geo}}|=64$ geometry tokens, and $|\mathbf{Q}^{\mathrm{act}}|=8$ action tokens. 
We set $\lambda_{\mathrm{act}}=\lambda_{\mathrm{rgb}}=\lambda_{\mathrm{rgb\text{-}act}}=\lambda_{\mathrm{geo\text{-}act}}=1.0$, and $\lambda_{\mathrm{geo}}=\lambda_{\mathrm{aff}}=0.1$, with $\lambda_{\mathrm{anc}}=1.0$ for both the geometry and affordance losses and $\lambda_{\mathrm{dice}}=1.0$. 

For supervised fine-tuning (SFT), we optimize the model for 14,140 steps with a global batch size of 256 using AdamW, a peak learning rate of $5\times10^{-5}$, 404 warmup steps, and cosine decay. Following SFT, we further optimize the action diffusion model through reinforcement learning (RL) with GRPO \citep{shao2024deepseekmath} for 4,500 steps with a learning rate of $3\times10^{-6}$ after 50 warmup steps, while keeping the rest of the model frozen. Detailed hyperparameters and training configurations for both SFT and RL, including the RL reward and rollout settings, are provided in Appendix~\ref{app:training}.

\subsection{Main Results}
\paragraph{NAVSIM v1 Results.}
We evaluate AffordDrive3D on the NAVSIM v1 benchmark against end-to-end, VLA and WAM baselines \citep{yuan2024drama, li2024hydra, liao2025diffusiondrive, zhou2025autovla, xiong2026recogdrive, li2025end, zhang2026resworld, chen2025drivinggpt, zhang2025epona, zhao2026forecasting, li2026drivevla, zhou2026drivedreamer, zeng2026futuresightdrive, sheng2026explorevla, cheng2026owmdrive}. In Table~\ref{navsim}, AffordDrive3D achieves a PDMS of 91.3, outperforming all baselines in the comparison. In particular, it improves over ExploreVLA \citep{sheng2026explorevla}, which also leverages RGB and depth supervision, by 0.9 PDMS. AffordDrive3D also achieves a strong DAC score of 99.0, together with NC and TTC scores of 98.6 and 96.3, while maintaining an EP score of 84.7. These results suggest that jointly modeling future driving affordances and geometric structure provides useful action-relevant spatial supervision for trajectory planning.

\begin{table}[t]
  \vspace{-0.3cm}
  \caption{Comparison with baselines on the NAVSIM v1 benchmark. Best results are \textbf{bolded}.}
  \label{navsim}
  \centering
  \small
  \setlength{\tabcolsep}{5pt}
  \renewcommand{\arraystretch}{1.15}
  \begin{tabular}{l|c|cccccc}
    \toprule
    Method & World Sup. & NC $\uparrow$ & DAC $\uparrow$   & TTC $\uparrow$ & Comf. $\uparrow$ & EP $\uparrow$ & PDMS $\uparrow$  \\
    \midrule
    DRAMA & - &  98.0 & 93.1 & 94.8 & \textbf{100.0} & 80.1 & 85.5 \\
    Hydra-MDP & - & 98.3 & 96.0 & 94.6 & \textbf{100.0} & 78.7 & 86.5\\
    DiffusionDrive & - & 98.2 & 96.2 & 94.7 & \textbf{100.0} & 82.2 & 88.1 \\
    AutoVLA & - & 98.4 & 95.6 & \textbf{98.0} & 99.9 & 81.9 & 89.1 \\
    ReCogDrive & - &  97.9 & 97.3 & 94.9 & \textbf{100.0} & \textbf{87.3} & 90.8 \\
    WoTE & BEV & 98.5 & 96.8 & 94.9 &99.9 &81.9 & 88.3 \\
    ResWorld & BEV & \textbf{98.9} & 96.5 & 95.6 & \textbf{100.0} & 83.1 & 89.0\\
    DrivingGPT & RGB & \textbf{98.9} & 90.7 & 94.9 & 95.6 & 79.7 & 82.4\\
    Epona & RGB & 97.9 & 95.1 & 93.8 & 99.9 & 80.4 & 86.2\\
    PWM & RGB & 98.6 & 95.9 & 95.4 & \textbf{100.0} & 81.8 & 88.1 \\
    DriveVLA-W0 & RGB & 98.7 & \textbf{99.1} & 95.3 & 99.3 & 83.3 & 90.2\\
    DriveDreamer-Policy & RGB + Current Depth & 98.4 & 97.1 & 95.1& \textbf{100.0} & 83.5 & 89.2\\
    FSDrive & RGB + CoT & 98.2 & 93.8 & 93.3 & 99.9 & 80.1 & 85.1\\
    ExploreVLA & RGB + Depth & 98.8 & 98.4 & 96.5 & 99.9 & 83.5 & 90.4\\
    
    OWMDrive & 3D Occupancy & 98.2 & 97.6 & 94.6 & 99.8 & \textbf{87.3} & 90.8 \\

    \midrule
    Ours & Aff + RGB + Depth & 98.6 & 99.0 & 96.3 & \textbf{100.0} & 84.7 & \textbf{91.3} \\
  
    \bottomrule
  \end{tabular}
\end{table}

\paragraph{NAVSIM v2 Results.}

We further compare AffordDrive3D with baseline methods \citep{zou2025diffusiondrivev2, yao2026drivesuprim, li2026drivevla, zhou2026drivedreamer, sheng2026explorevla} on NAVSIM v2, which extends the evaluation to a broader set of driving metrics. As shown in \cref{tab:navsim-main-results}, AffordDrive3D achieves the highest EPDMS of 89.9, improving over ExploreVLA \citep{sheng2026explorevla} and DriveDreamer-Policy \citep{zhou2026drivedreamer} by 1.1 and 1.2 points, respectively. These results further demonstrate the strong planning performance of AffordDrive3D and the effectiveness of jointly modeling future driving affordances and geometric structure.

\begin{table}[t]
  \centering
  \caption{Comparison with baselines on NAVSIM v2 with extended metrics. Best results are \textbf{bolded}.}
  \label{tab:navsim-main-results}
  \small
  \setlength{\tabcolsep}{4pt}
  \renewcommand{\arraystretch}{1.15}
  \begin{tabular}{@{}l|cccc|ccccc|c@{}}
    \toprule
    \textbf{Model}
      & NC $\uparrow$
      & DAC $\uparrow$
      & DDC $\uparrow$
      & TLC $\uparrow$
      & EP $\uparrow$
      & TTC $\uparrow$
      & LK $\uparrow$
      & HC $\uparrow$
      & EC $\uparrow$
      & EPDMS $\uparrow$ \\
    \midrule
    DiffusionDriveV2 &97.7 & 96.6 & 99.2 & 99.8& 88.9 &  97.2 &  96.0&97.8 & \textbf{91.0} & 85.5\\
    DriveSuprim & 98.4  & 98.6 & 99.6& 99.8& \textbf{90.5} & 97.8& 97.0& \textbf{98.3}& 78.6& 87.1 \\
    DriveVLA-W0 & 98.5&  \textbf{99.1}  & 98.0 &  99.7 & 86.4 &98.1 &93.2  &  97.9&58.9 &86.1 \\
    DriveDreamer-Policy & 98.4 & 97.1 & 99.5& \textbf{99.9}& 87.9 & 97.7& 97.6& \textbf{98.3}& 79.4& 88.7\\
    ExploreVLA & \textbf{98.8} &96.2 & 99.6 & 99.8 & 87.1 &  \textbf{98.2}& \textbf{97.8} & \textbf{98.3} & 86.8 & 88.8 \\
    \midrule
    Ours &  98.4 & 98.2 & \textbf{99.7} & \textbf{99.9} & 87.9 & 97.8& 97.4& \textbf{98.3}& 80.4& \textbf{89.9}\\
    \bottomrule
  \end{tabular}
  \vspace{-0.3cm}
\end{table}

\begin{table}[t]
  \centering
  \caption{Ablation study of affordance, RGB, and geometry future modeling. For the \emph{+ Geometry} setting, the RGB branch provides latent context for depth prediction, but its gradients are stopped from propagating to the shared backbone. Results are reported without RL training.}
  \label{tab:ablation}
  \setlength{\tabcolsep}{4pt}
  \begin{tabular}{@{}lccccccccc@{}}
    \toprule
    Setting
      & Aff.
      & RGB
      & Geo.
      & NC $\uparrow$
      & DAC $\uparrow$
      & TTC $\uparrow$
      & Comf. $\uparrow$
      & EP $\uparrow$
      & PDMS $\uparrow$ \\
    \midrule
    Action only
      &--  &  --&-- & 97.9 & 96.4 & 94.3
      & \textbf{100.0} & 82.6 & 88.1 \\
    + Affordance
      & $\checkmark$ & -- & --
      & 98.3 & 97.1 & 95.1 & \textbf{100.0} & 83.5 & 89.2 \\
    + RGB
      & -- & $\checkmark$ & --
      & 98.3 & 96.8 & \textbf{95.3} & \textbf{100.0} & 83.0 & 88.9 \\
    + Geometry
      & -- & -- & $\checkmark$
      & 98.2 & 97.1 & 95.2 & \textbf{100.0} & 83.2 & 89.1 \\
    \addlinespace
    Full model
      & $\checkmark$ & $\checkmark$ & $\checkmark$
      & \textbf{98.4} & \textbf{97.6} &  95.2 & \textbf{100.0} & \textbf{83.7} & \textbf{89.6} \\
    \bottomrule
  \end{tabular}
  \vspace{-0.3cm}
\end{table}

\begin{figure}[t]
    \centering
    \includegraphics[width=0.99\linewidth]{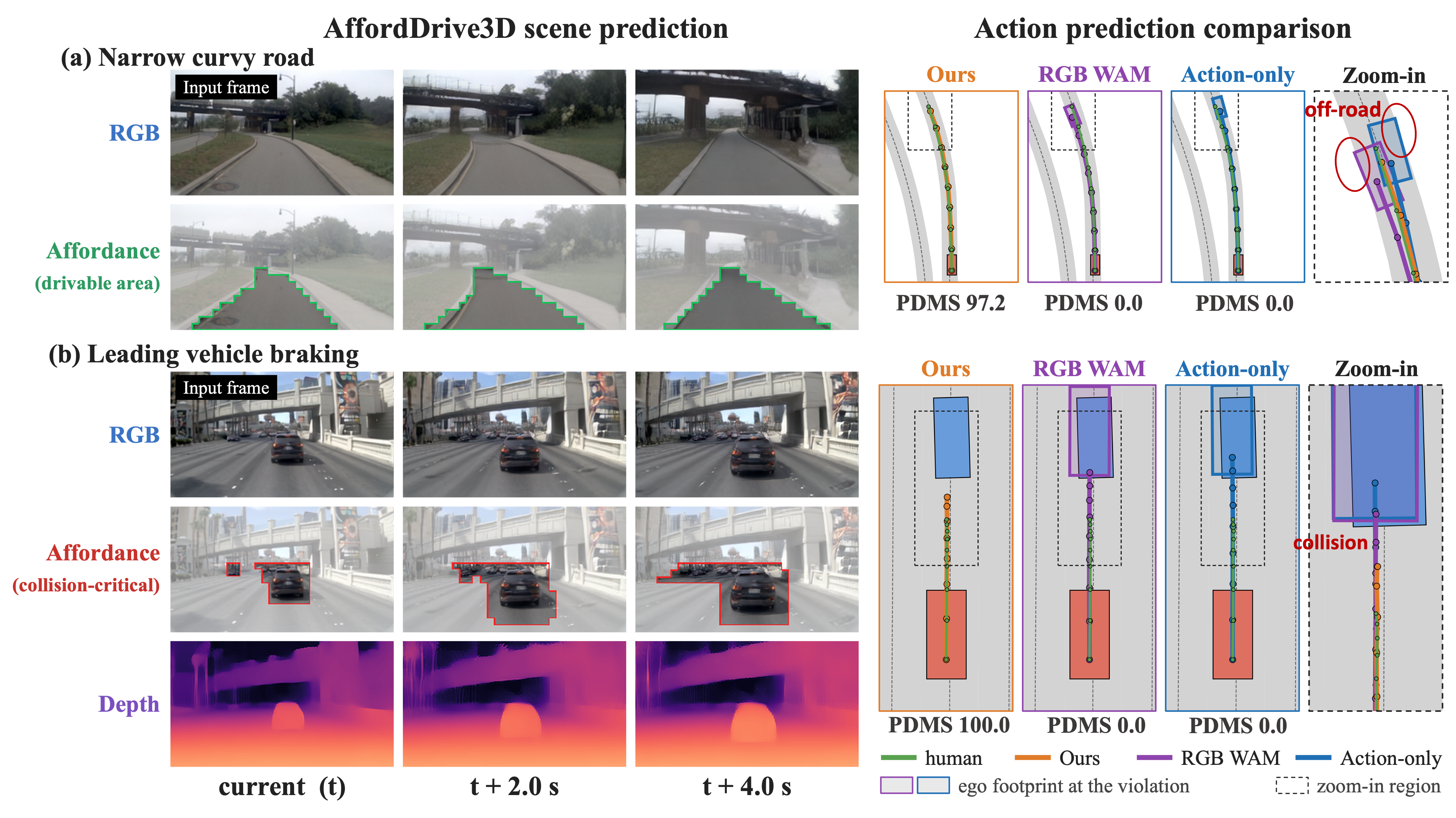}
    \caption{\textbf{Qualitative results of AffordDrive3D in two challenging scenarios.} (a) The narrow curvy-road scenario: AffordDrive3D forecasts the road structure and remains on-road, while 
    both baselines deviate from the lane. (b) The leading-vehicle braking scenario: AffordDrive3D captures the potential collision risk from the leading vehicle together with RGB-consistent future geometry, maintaining a safe trajectory while both baselines collide. In all BEV visualizations, highlighted boxes show the ego footprint at the violation state obtained from the NAVSIM rollout, while other-agent boxes show their positions at $t=0$.}
    \label{fig:qualitative}
    \vspace{-0.4cm}
\end{figure}

\begin{table}[t]
\centering

\begin{minipage}[t]{0.55\linewidth}
  \centering
  \caption{Affordance prediction across horizons.}
  \label{tab:affordance-horizons}
  \small
  \setlength{\tabcolsep}{4pt}
  \begin{tabular}{@{}ccc@{}}
    \toprule
    $t$ & $\mathrm{AP}_{\text{Collision}}$ & $\mathrm{AP}_{\text{Drivable Area}}$ \\
    \midrule
    Current             & 0.918 & 0.986 \\
    $0.5$--$2$ s (avg.)   & 0.846 & 0.983 \\
    $2.5$--$4$ s (avg.)   & 0.732 & 0.971 \\
    \bottomrule
  \end{tabular}
\end{minipage}
\hfill
\begin{minipage}[t]{0.44\linewidth}
  \centering
  \caption{Current-frame affordance prediction using frozen and unfrozen VLM.}
  \label{tab:affordance-mlp}
  \small
  \setlength{\tabcolsep}{4pt}
  \begin{tabular}{@{}ccc@{}}
    \toprule
     & $\mathrm{AP}_{\text{Collision}}$ & $\mathrm{AP}_{\text{Drivable Area}}$ \\
    \midrule
    Frozen VLM   & 0.782 & 0.931 \\
    Unfrozen VLM & \textbf{0.912} & \textbf{0.985} \\
    \bottomrule
  \end{tabular}
\end{minipage}
\vspace{-0.6cm}
\end{table}

\paragraph{Qualitative Results.}

\Cref{fig:qualitative} visualizes the current and future RGB, affordance, and depth predictions of AffordDrive3D, together with trajectory predictions from an action-only baseline and the RGB-based world-action model DriveVLA-W0 \citep{li2026drivevla}. In the narrow curvy-road example in \cref{fig:qualitative}(a), both baselines deviate from the drivable region and go off-road, whereas AffordDrive3D remains near the lane center. The predicted drivable-area maps consistently capture the narrow drivable road structure over time, illustrating how explicit affordance forecasting can provide important cues about where the ego vehicle can safely move. In the leading-vehicle braking example in \cref{fig:qualitative}(b), the two baselines collide with the vehicle ahead, while AffordDrive3D maintains a safe distance. Its collision-critical prediction correctly highlights the leading vehicle across future horizons, while the depth prediction, consistent with the predicted RGB, captures the leading vehicle’s relative distance and surrounding scene geometry. These examples illustrate how jointly modeling driving affordances and geometry can provide action-relevant spatial information for challenging planning scenarios. Additional qualitative comparisons and visualizations are provided in Appendix~\ref{app:more_vis}. 

\subsection{Ablation Studies}

\paragraph{Ablation of Affordance and World Modeling.}
We perform an ablation study to understand how affordance, RGB, and geometry modeling affect planning performance in \cref{tab:ablation}. Since our depth prediction is conditioned on the denoised RGB latents, the geometry-only setting still retains RGB prediction to provide the required latent input. However, we stop the RGB diffusion gradients from propagating back to the shared backbone, so the RGB branch only provides the latent context required by the geometry branch and does not directly affect the shared representation used for action prediction. Compared to the action-only baseline at 88.1 PDMS, adding affordance, RGB, or geometry improves performance to 89.2, 88.9, and 89.1 PDMS, respectively. The full model achieves the best performance of 89.6 PDMS, with the highest NC, DAC, and EP, demonstrating the effectiveness of jointly modeling RGB, geometry, and affordance.

\paragraph{Temporal Affordance Evaluation.}
We evaluate the predicted driving affordances across different future horizons in \cref{tab:affordance-horizons}. We report AP averaged over the near-term prediction horizons (0.5, 1.0, 1.5, and 2.0\,s) and the longer-term prediction horizons (2.5, 3.0, 3.5, and 4.0\,s). AffordDrive3D achieves strong current-frame prediction, with an AP of 0.918 for collision-critical regions and 0.986 for drivable areas. Drivable-area prediction remains highly stable over time, reaching 0.983 AP over the near-term horizon and 0.971 over the longer-term horizon. Compared with drivable-area prediction, collision prediction degrades more noticeably at longer horizons, with AP decreasing from 0.846 to 0.732. This is expected because drivable areas are largely determined by relatively stable road structure, whereas future collision-critical regions depend on anticipating the motion of dynamic agents such as vehicles and pedestrians. Overall, the results show that AffordDrive3D can reliably forecast both where the ego vehicle can safely move and where potential collision risks may emerge over time.

\paragraph{Affordance Information in the VLM Representation.}
We adopt a VLM backbone to provide the contextual scene understanding needed for affordance prediction. To assess whether the original VLM representation already contains information useful for identifying regions relevant to ego motion, we remove all world-modeling and action components and attach two lightweight MLP heads directly to the VLM visual tokens to predict current-frame drivable-area and collision-critical maps. We compare a frozen setting, where only the prediction heads are trained, with an unfrozen setting, where the VLM backbone is jointly optimized with the affordance objectives. As shown in \cref{tab:affordance-mlp}, the frozen setting achieves APs of 0.782 and 0.931 for collision-critical and drivable-area prediction, respectively. In the unfrozen setting, the APs increase to 0.912 and 0.985, with a particularly large gain for collision-critical prediction. These results indicate that the original VLM representation already contains useful information for identifying driving-relevant regions, while explicit affordance supervision further enriches the backbone with affordance-relevant information.

\section{Conclusion}

We presented \textbf{AffordDrive3D}, an affordance- and geometry-aware world-action model for autonomous driving. Beyond RGB-based future prediction, AffordDrive3D models future geometry to provide explicit spatial structure and forecasts temporal driving affordances, including drivable and collision-critical regions, to support trajectory planning. By jointly learning future appearance, geometry, affordance, and action within a shared VLM backbone, AffordDrive3D more closely aligns future world modeling with the planning objective. AffordDrive3D achieves 91.3 PDMS and 89.9 EPDMS on NAVSIM, while ablation and qualitative results demonstrate the effectiveness of future geometry modeling and temporal affordance forecasting for trajectory planning.

\bibliography{references}

@inproceedings{sima2024drivelm,
  title={Drivelm: Driving with graph visual question answering},
  author={Sima, Chonghao and Renz, Katrin and Chitta, Kashyap and Chen, Li and Zhang, Hanxue and Xie, Chengen and Bei{\ss}wenger, Jens and Luo, Ping and Geiger, Andreas and Li, Hongyang},
  booktitle={European conference on computer vision},
  pages={256--274},
  year={2024},
  organization={Springer}
}

@inproceedings{tian2024drivevlm,
title={Drive{VLM}: The Convergence of Autonomous Driving and Large Vision-Language Models},
author={Xiaoyu Tian and Junru Gu and Bailin Li and Yicheng Liu and Yang Wang and Zhiyong Zhao and Kun Zhan and Peng Jia and XianPeng Lang and Hang Zhao},
booktitle={8th Annual Conference on Robot Learning},
year={2024},
url={https://openreview.net/forum?id=928V4Umlys}
}

@article{zhou2025autovla,
 title={AutoVLA: A Vision-Language-Action Model for End-to-End Autonomous Driving with Adaptive Reasoning and Reinforcement Fine-Tuning},
 author={Zhou, Zewei and Cai, Tianhui and Zhao, Seth Z. and Zhang, Yun and Huang, Zhiyu and Zhou, Bolei and Ma, Jiaqi},
 journal={Advances in Neural Information Processing Systems (NeurIPS)},
 year={2025}
}

@inproceedings{bassole2026hybriddrivevla,
  title={HybridDriveVLA: Vision-Language-Action Model with Visual CoT reasoning and ToT Evaluation for Autonomous Driving},
  author={Bassole, Yipene Cedric Francois and Kim, Sungwoo and Jung, Jiwoo and Sung, Yunsick},
  booktitle={Proceedings of the IEEE/CVF Conference on Computer Vision and Pattern Recognition},
  pages={32421--32430},
  year={2026}
}

@inproceedings{peng2026colavla,
  title={Colavla: Leveraging cognitive latent reasoning for hierarchical parallel trajectory planning in autonomous driving},
  author={Peng, Qihang and Chen, Xuesong and Yang, Chenye and Shi, Shaoshuai and Li, Hongsheng},
  booktitle={Proceedings of the IEEE/CVF Conference on Computer Vision and Pattern Recognition},
  pages={17809--17819},
  year={2026}
}

@inproceedings{wang2024driving,
  title={Driving into the future: Multiview visual forecasting and planning with world model for autonomous driving},
  author={Wang, Yuqi and He, Jiawei and Fan, Lue and Li, Hongxin and Chen, Yuntao and Zhang, Zhaoxiang},
  booktitle={2024 IEEE/CVF Conference on Computer Vision and Pattern Recognition (CVPR)},
  pages={14749--14759},
  year={2024},
  organization={IEEE}
}

@inproceedings{yang2025driving,
  title={Driving in the occupancy world: Vision-centric 4d occupancy forecasting and planning via world models for autonomous driving},
  author={Yang, Yu and Mei, Jianbiao and Ma, Yukai and Du, Siliang and Chen, Wenqing and Qian, Yijie and Feng, Yuxiang and Liu, Yong},
  booktitle={Proceedings of the AAAI Conference on Artificial Intelligence},
  volume={39},
  number={9},
  pages={9327--9335},
  year={2025}
}

@article{shi2026drivewam,
  title={DriveWAM: Video Generative Priors Enable Scalable World-Action Modeling for Autonomous Driving},
  author={Shi, Chen and Xu, Jinrui and Shi, Shaoshuai and Sheng, Kehua and Zhang, Bo and Jiang, Li},
  journal={arXiv preprint arXiv:2605.28544},
  year={2026}
}

@article{wang2026wa,
  title={WA-JEPA: Rethinking the Video JEPA Paradigm for World-Action Modeling in Autonomous Driving},
  author={Wang, Xinlin and Xiang, Yujiao and Zhou, Yuheng and Wang, Jingqi and Huang, Minqing and Huang, Jiajie and Wei, Dongxu and Zhou, Tingguang and Wang, Xiyang and Chen, Gong and others},
  journal={arXiv preprint arXiv:2608.20974},
  year={2026}
}

@article{zhao2026simwam,
  title={SimWAM: A Simple World Action Model for End-to-End Autonomous Driving},
  author={Zhao, Zongchuang and Zhou, Xin and Xu, Tianyang and Sun, Zhengyang and Zhou, Kaixuan and Li, Honglin and Liang, Dingkang and Bai, Xiang},
  journal={arXiv preprint arXiv:2608.07468},
  year={2026}
}

@article{wang2026sv,
  title={SV-WAM: An Efficient Surround-View World-Action Model for End-to-End Autonomous Driving},
  author={Wang, Jinyang and Li, Shiwei and Wang, Junjian and Deng, Zhiqiang and Gao, Jianbin and Zhao, Yihang and Liu, Liu and Zhao, Yongjia and Chen, Jinlong and Xu, Huirui and others},
  journal={arXiv preprint arXiv:2609.03602},
  year={2026}
}

@inproceedings{li2026spacedrive,
  title={Spacedrive: Infusing spatial awareness into vlm-based autonomous driving},
  author={Li, Peizheng and Zhang, Zhenghao and Holtz, David and Yu, Hang and Yang, Yutong and Lai, Yuzhi and Song, Rui and Geiger, Andreas and Zell, Andreas},
  booktitle={Proceedings of the IEEE/CVF Conference on Computer Vision and Pattern Recognition},
  pages={40096--40107},
  year={2026}
}

@article{lu2026geowam,
  title={GeoWAM: Visual Geometry World Action Models for Autonomous Driving},
  author={Lu, Yiren and Ye, Xin and Liu, Jiaming and Jacobson, Philip and Yao, Jin and Chen, Yi-chung and Merino, Liam and Kurra, Dhruva Dixith and Cai, Min and Lampo, Tom and others},
  journal={arXiv preprint arXiv:2608.23486},
  year={2026}
}

@article{yu2026affordancevla,
  title={AffordanceVLA: A Vision-Language-Action Model Empowering Action Generation through Affordance-Aware Understanding},
  author={Yu, Qize and You, Jiadi and Wang, Yuran and Liang, Jiaqi and Ping, Bowen and Tian, Yang and Chen, Yue and Cai, Minghong and Gong, Zeying and Wu, Ruihai and others},
  journal={arXiv preprint arXiv:2606.06155},
  year={2026}
}

@inproceedings{hu2023planning,
  title={Planning-oriented autonomous driving},
  author={Hu, Yihan and Yang, Jiazhi and Chen, Li and Li, Keyu and Sima, Chonghao and Zhu, Xizhou and Chai, Siqi and Du, Senyao and Lin, Tianwei and Wang, Wenhai and others},
  booktitle={2023 IEEE/CVF Conference on Computer Vision and Pattern Recognition (CVPR)},
  pages={17853--17862},
  year={2023},
  organization={IEEE}
}

@inproceedings{jiang2023vad,
  title={Vad: Vectorized scene representation for efficient autonomous driving},
  author={Jiang, Bo and Chen, Shaoyu and Xu, Qing and Liao, Bencheng and Chen, Jiajie and Zhou, Helong and Zhang, Qian and Liu, Wenyu and Huang, Chang and Wang, Xinggang},
  booktitle={2023 IEEE/CVF International Conference on Computer Vision (ICCV)},
  pages={8306--8316},
  year={2023},
  organization={IEEE}
}

@inproceedings{yang2026geniedrive,
  title={Geniedrive: Towards physics-aware driving world model with 4d occupancy guided video generation},
  author={Yang, Zhenya and Liu, Zhe and Lu, Yuxiang and Hou, Liping and Miao, Chenxuan and Peng, Siyi and Feng, Bailan and Bai, Xiang and Zhao, Hengshuang},
  booktitle={Proceedings of the IEEE/CVF Conference on Computer Vision and Pattern Recognition},
  pages={35680--35690},
  year={2026}
}

@article{wang2026afford,
  title={Afford-VLA: Action-Aligned Visual Planning via Internalized Affordance},
  author={Wang, Runze and Fu, Yuqian and Li, Yu and Lin, Tao and Qian, Tianwen and Elhoseiny, Mohamed and Zhao, Bo and Fu, Yanwei and Jiang, Yu-Gang and Xue, Xiangyang},
  journal={arXiv preprint arXiv:2605.24203},
  year={2026}
}

@article{zhou2026drivedreamer,
  title={Drivedreamer-policy: A geometry-grounded world-action model for unified generation and planning},
  author={Zhou, Yang and Wang, Xiaofeng and Shao, Hao and Wang, Letian and Zhao, Guosheng and Shao, Jiangnan and Zhu, Jiagang and Yu, Tingdong and Zhu, Zheng and Huang, Guan and others},
  journal={arXiv preprint arXiv:2604.01765},
  year={2026}
}

@article{dauner2024navsim,
  title={Navsim: Data-driven non-reactive autonomous vehicle simulation and benchmarking},
  author={Dauner, Daniel and Hallgarten, Marcel and Li, Tianyu and Weng, Xinshuo and Huang, Zhiyu and Yang, Zetong and Li, Hongyang and Gilitschenski, Igor and Ivanovic, Boris and Pavone, Marco and others},
  journal={Advances in Neural Information Processing Systems},
  volume={37},
  pages={28706--28719},
  year={2024}
}

@inproceedings{liao2025diffusiondrive,
  title={Diffusiondrive: Truncated diffusion model for end-to-end autonomous driving},
  author={Liao, Bencheng and Chen, Shaoyu and Yin, Haoran and Jiang, Bo and Wang, Cheng and Yan, Sixu and Zhang, Xinbang and Li, Xiangyu and Zhang, Ying and Zhang, Qian and others},
  booktitle={2025 IEEE/CVF Conference on Computer Vision and Pattern Recognition (CVPR)},
  pages={12037--12047},
  year={2025},
  organization={IEEE}
}

@inproceedings{zhou2026opendrivevla,
  title={Opendrivevla: Towards end-to-end autonomous driving with large vision language action model},
  author={Zhou, Xingcheng and Han, Xuyuan and Yang, Feng and Ma, Yunpu and Tresp, Volker and Knoll, Alois},
  booktitle={Proceedings of the AAAI Conference on Artificial Intelligence},
  volume={40},
  number={16},
  pages={13782--13790},
  year={2026}
}

@inproceedings{renz2025simlingo,
  title={Simlingo: Vision-only closed-loop autonomous driving with language-action alignment},
  author={Renz, Katrin and Chen, Long and Arani, Elahe and Sinavski, Oleg},
  booktitle={2025 IEEE/CVF Conference on Computer Vision and Pattern Recognition (CVPR)},
  pages={11993--12003},
  year={2025},
  organization={IEEE}
}

@inproceedings{fu2025orion,
  title={Orion: A holistic end-to-end autonomous driving framework by vision-language instructed action generation},
  author={Fu, Haoyu and Zhang, Diankun and Zhao, Zongchuang and Cui, Jianfeng and Liang, Dingkang and Zhang, Chong and Zhang, Dingyuan and Xie, Hongwei and Wang, Bing and Bai, Xiang},
  booktitle={2025 IEEE/CVF International Conference on Computer Vision (ICCV)},
  pages={24823--24834},
  year={2025},
  organization={IEEE}
}

@article{wang2026learning,
  title={Learning Vision-Language-Action World Models for Autonomous Driving},
  author={Wang, Guoqing and Tang, Pin and Ren, Xiangxuan and Zhao, Guodongfang and Feng, Bailan and Ma, Chao},
  journal={arXiv preprint arXiv:2604.09059},
  year={2026}
}

@article{liu2026driveworld,
  title={Driveworld-vla: Unified latent-space world modeling with vision-language-action for autonomous driving},
  author={Liu, Lin and Song, Ziying and Jia, Caiyan and Ye, Hangjun and Hao, Xiaoshuai and Chen, Long and others},
  journal={arXiv preprint arXiv:2602.06521},
  year={2026}
}

@inproceedings{li2026drivevla,
  title={Drivevla-w0: World models amplify data scaling law in autonomous driving},
  author={Li, Yingyan and Shang, Shuyao and Liu, Weisong and Zhan, Bing and Wang, Haochen and Wang, Yuqi and Chen, Yuntao and Wang, Xiaoman and An, Yasong and Tang, Chufeng and others},
  booktitle={International Conference on Learning Representations},
  volume={2026},
  pages={7890--7911},
  year={2026}
}

@article{zhao2026forecasting,
  title={From forecasting to planning: Policy world model for collaborative state-action prediction},
  author={Zhao, Zhida and Fu, Talas and Wang, Yifan and Wang, Lijun and Lu, Huchuan},
  journal={Advances in Neural Information Processing Systems},
  volume={38},
  pages={134585--134611},
  year={2026}
}

@article{zhang2026resworld,
  title={Resworld: Temporal residual world model for end-to-end autonomous driving},
  author={Zhang, Jinqing and Fu, Zehua and Xu, Zelin and Dai, Wenying and Liu, Qingjie and Wang, Yunhong},
  journal={arXiv preprint arXiv:2602.10884},
  year={2026}
}

@inproceedings{chen2025drivinggpt,
  title={Drivinggpt: Unifying driving world modeling and planning with multi-modal autoregressive transformers},
  author={Chen, Yuntao and Wang, Yuqi and Zhang, Zhaoxiang},
  booktitle={2025 IEEE/CVF International Conference on Computer Vision (ICCV)},
  pages={26890--26900},
  year={2025},
  organization={IEEE}
}

@article{hong2026drivefuture,
  title={DriveFuture: Future-Aware Latent World Models for Autonomous Driving},
  author={Hong, Yufeng and Zhou, Xiaotian and Li, Yingyan and Zhou, Xiangpo and Liu, Lin and Luo, Yadan and Xu, Shaoqing and Yang, Lei and Song, Ziying},
  journal={arXiv preprint arXiv:2605.09701},
  year={2026}
}

@inproceedings{li2025end,
  title={End-to-end driving with online trajectory evaluation via bev world model},
  author={Li, Yingyan and Wang, Yuqi and Liu, Yang and He, Jiawei and Fan, Lue and Zhang, Zhaoxiang},
  booktitle={2025 IEEE/CVF International Conference on Computer Vision (ICCV)},
  pages={27137--27146},
  year={2025},
  organization={IEEE}
}

@article{zhang2026geoworldad,
  title={GeoWorldAD: Geometry World Action Model for Autonomous Driving},
  author={Zhang, Songyan and Tian, Jinyuan and Li, Hanbing and Liu, Daqi and Chen, Hao and Huang, Wenhui and Li, Fang and Chen, Guang and Ye, Hangjun and Chen, Long and others},
  journal={arXiv preprint arXiv:2607.17521},
  year={2026}
}

@inproceedings{li2025enhancing,
  title={Enhancing end-to-end autonomous driving with latent world model},
  author={Li, Yingyan and Fan, Lue and He, Jiawei and Wang, Yuqi and Chen, Yuntao and Zhang, Zhaoxiang and Tan, Tieniu},
  booktitle={International Conference on Learning Representations},
  volume={2025},
  pages={42942--42959},
  year={2025}
}

@article{cheng2026owmdrive,
  title={OWMDrive: Causality-Aware End-to-End Autonomous Driving via 4D Occupancy World Model},
  author={Cheng, Junjie and Song, Ruiqi and Wu, Ye and Zeng, Nanxing and Li, Ximiao and Ai, Yunfeng},
  journal={arXiv preprint arXiv:2606.30421},
  year={2026}
}

@inproceedings{sheng2026explorevla,
  title={Explorevla: Dense world modeling and exploration for end-to-end autonomous driving},
  author={Sheng, Zihao and Ye, Xin and Luo, Jingru and Chen, Sikai and Ren, Liu},
  booktitle={European Conference on Computer Vision},
  pages={266--284},
  year={2026},
  organization={Springer}
}

@article{bai2025qwen3,
  title={Qwen3-vl technical report},
  author={Bai, Shuai and Cai, Yuxuan and Chen, Ruizhe and Chen, Keqin and Chen, Xionghui and Cheng, Zesen and Deng, Lianghao and Ding, Wei and Gao, Chang and Ge, Chunjiang and others},
  journal={arXiv preprint arXiv:2511.21631},
  year={2025}
}

@article{wan2025,
      title={Wan: Open and Advanced Large-Scale Video Generative Models}, 
      author={Team Wan and Ang Wang and Baole Ai and Bin Wen and Chaojie Mao and Chen-Wei Xie and Di Chen and Feiwu Yu and Haiming Zhao and Jianxiao Yang and Jianyuan Zeng and Jiayu Wang and Jingfeng Zhang and Jingren Zhou and Jinkai Wang and Jixuan Chen and Kai Zhu and Kang Zhao and Keyu Yan and Lianghua Huang and Mengyang Feng and Ningyi Zhang and Pandeng Li and Pingyu Wu and Ruihang Chu and Ruili Feng and Shiwei Zhang and Siyang Sun and Tao Fang and Tianxing Wang and Tianyi Gui and Tingyu Weng and Tong Shen and Wei Lin and Wei Wang and Wei Wang and Wenmeng Zhou and Wente Wang and Wenting Shen and Wenyuan Yu and Xianzhong Shi and Xiaoming Huang and Xin Xu and Yan Kou and Yangyu Lv and Yifei Li and Yijing Liu and Yiming Wang and Yingya Zhang and Yitong Huang and Yong Li and You Wu and Yu Liu and Yulin Pan and Yun Zheng and Yuntao Hong and Yupeng Shi and Yutong Feng and Zeyinzi Jiang and Zhen Han and Zhi-Fan Wu and Ziyu Liu},
      journal = {arXiv preprint arXiv:2503.20314},
      year={2025}
}

@article{li2025generalized,
  title={Generalized trajectory scoring for end-to-end multimodal planning},
  author={Li, Zhenxin and Yao, Wenhao and Wang, Zi and Sun, Xinglong and Chen, Joshua and Chang, Nadine and Shen, Maying and Wu, Zuxuan and Lan, Shiyi and Alvarez, Jose M},
  journal={arXiv preprint arXiv:2506.06664},
  year={2025}
}

@article{li2025ztrs,
  title={Ztrs: Zero-imitation end-to-end autonomous driving with trajectory scoring},
  author={Li, Zhenxin and Yao, Wenhao and Wang, Zi and Sun, Xinglong and Chen, Jingde and Chang, Nadine and Shen, Maying and Song, Jingyu and Wu, Zuxuan and Lan, Shiyi and others},
  journal={arXiv preprint arXiv:2510.24108},
  year={2025}
}

@article{yao2026had,
  title={HAD: Combining Hierarchical Diffusion with Metric-Decoupled RL for End-to-End Driving},
  author={Yao, Wenhao and Sun, Xinglong and Li, Zhenxin and Lan, Shiyi and Wang, Zi and Alvarez, Jose M and Wu, Zuxuan},
  journal={arXiv preprint arXiv:2604.03581},
  year={2026}
}

@article{li2026large,
  title={Large Discrete Policy: Advancing Explicit Behavior Modeling with Stochastic Iterative Scoring},
  author={Li, Zhenxin and Chang, Nadine and Sun, Xinglong and Chen, Jingde and Yao, Wenhao and Wang, Zi and Shen, Maying and Jiang, Yu-Gang and Wu, Zuxuan and Lan, Shiyi and others},
  journal={arXiv preprint arXiv:2609.07049},
  year={2026}
}

@article{liang2025unifuture,
  title={Unifuture: A 4d driving world model for future generation and perception},
  author={Liang, Dingkang and Zhang, Dingyuan and Zhou, Xin and Tu, Sifan and Feng, Tianrui and Li, Xiaofan and Zhang, Yumeng and Du, Mingyang and Tan, Xiao and Bai, Xiang},
  journal={arXiv preprint arXiv:2503.13587},
  year={2025}
}

@article{zeng2026futuresightdrive,
  title={Futuresightdrive: Thinking visually with spatio-temporal cot for autonomous driving},
  author={Zeng, Shuang and Chang, Xinyuan and Xie, Mengwei and Liu, Xinran and Bai, Yifan and Pan, Zheng and Xu, Mu and Wei, Xing},
  journal={Advances in Neural Information Processing Systems},
  volume={38},
  pages={67299--67318},
  year={2026}
}

@article{li2024hydra,
  title={Hydra-mdp: End-to-end multimodal planning with multi-target hydra-distillation},
  author={Li, Zhenxin and Li, Kailin and Wang, Shihao and Lan, Shiyi and Yu, Zhiding and Ji, Yishen and Li, Zhiqi and Zhu, Ziyue and Kautz, Jan and Wu, Zuxuan and others},
  journal={arXiv preprint arXiv:2406.06978},
  year={2024}
}

@article{yuan2024drama,
  title={Drama: An efficient end-to-end motion planner for autonomous driving with mamba},
  author={Yuan, Chengran and Zhang, Zhanqi and Sun, Jiawei and Sun, Shuo and Huang, Zefan and Lee, Christina Dao Wen and Li, Dongen and Han, Yuhang and Wong, Anthony and Tee, Keng Peng and others},
  journal={arXiv preprint arXiv:2408.03601},
  year={2024}
}

@inproceedings{xiong2026recogdrive,
  title={Recogdrive: A reinforced cognitive framework for end-to-end autonomous driving},
  author={Xiong, Kaixin and Guo, Xiangyu and Li, Fang and Yan, Sixu and Xu, Gangwei and Zhou, Lijun and Chen, Long and Sun, Haiyang and Wang, Bing and Ma, Kun and others},
  booktitle={International Conference on Learning Representations},
  volume={2026},
  pages={157518--157556},
  year={2026}
}

@inproceedings{zhang2025epona,
  title={Epona: Autoregressive diffusion world model for autonomous driving},
  author={Zhang, Kaiwen and Tang, Zhenyu and Hu, Xiaotao and Pan, Xingang and Guo, Xiaoyang and Liu, Yuan and Huang, Jingwei and Yuan, Li and Zhang, Qian and Long, Xiao-Xiao and others},
  booktitle={2025 IEEE/CVF International Conference on Computer Vision (ICCV)},
  pages={27220--27230},
  year={2025},
  organization={IEEE}
}

@article{zou2025diffusiondrivev2,
  title={Diffusiondrivev2: Reinforcement learning-constrained truncated diffusion modeling in end-to-end autonomous driving},
  author={Zou, Jialv and Chen, Shaoyu and Liao, Bencheng and Zheng, Zhiyu and Song, Yuehao and Zhang, Lefei and Zhang, Qian and Liu, Wenyu and Wang, Xinggang},
  journal={arXiv preprint arXiv:2512.07745},
  year={2025}
}

@inproceedings{yao2026drivesuprim,
  title={Drivesuprim: Towards precise trajectory selection for end-to-end planning},
  author={Yao, Wenhao and Li, Zhenxin and Lan, Shiyi and Wang, Zi and Sun, Xinglong and Alvarez, Jose M and Wu, Zuxuan},
  booktitle={Proceedings of the AAAI Conference on Artificial Intelligence},
  volume={40},
  number={14},
  pages={11910--11918},
  year={2026}
}

@article{depthanything3,
  title={Depth Anything 3: Recovering the visual space from any views},
  author={Haotong Lin and Sili Chen and Jun Hao Liew and Donny Y. Chen and Zhenyu Li and Guang Shi and Jiashi Feng and Bingyi Kang},
  journal={arXiv preprint arXiv:2511.10647},
  year={2025}
}

@article{shao2024deepseekmath,
  title={Deepseekmath: Pushing the limits of mathematical reasoning in open language models},
  author={Shao, Zhihong and Wang, Peiyi and Zhu, Qihao and Xu, Runxin and Song, Junxiao and Bi, Xiao and Zhang, Haowei and Zhang, Mingchuan and Li, YK and Wu, Yang and others},
  journal={arXiv preprint arXiv:2402.03300},
  year={2024}
}
\bibliographystyle{arxiv}

\appendix
\section*{Appendix}
\section{Affordance Label Construction}
\label{app:affordance}

\begin{figure}[t]
    \centering
    \includegraphics[width=\linewidth]{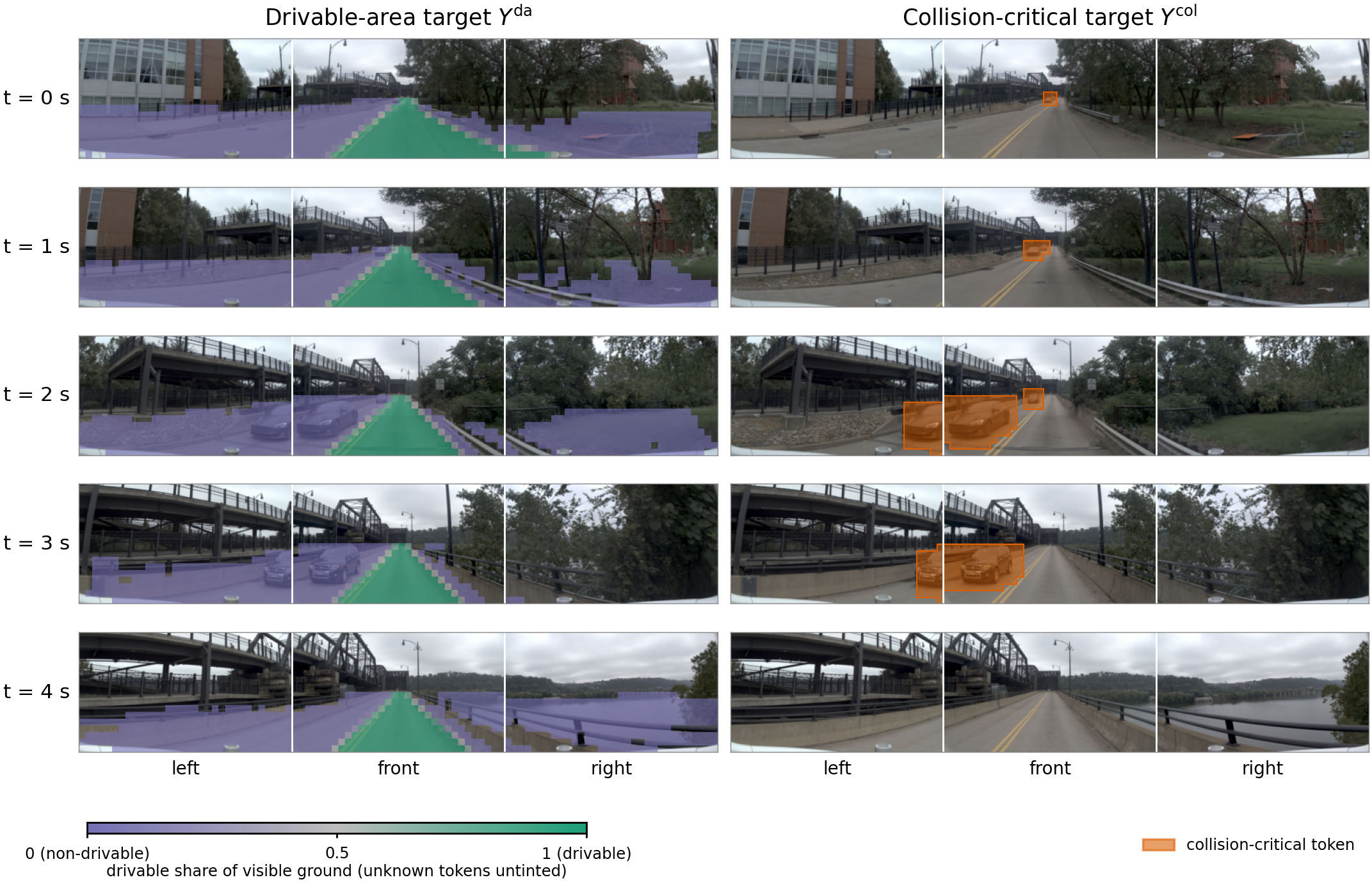}
    \caption{Example of ground-truth affordance targets at $t=0$--$4$\,s.}
    \label{fig:affordance_labels}
\end{figure}

We construct both labels directly on the visual-token grid used by the affordance heads. Each of the $C=3$ camera images is resized to $1024\times576$ and represented by an $18\times32$ grid of merged Qwen3-VL tokens, where each token corresponds to a $32\times32$ pixel patch. We generate labels at $t_k=0.5k$\,s for $k=0,\ldots,8$. \Cref{fig:affordance_labels} shows an example of affordance labels.

\paragraph{Drivable area.}
For each timestamp, we first identify the road surface that is lawfully reachable from the ego vehicle. We seed the NAVSIM roadblock graph with the roadblocks containing the ground-truth ego pose, follow directed outgoing connections for up to six hops, and retain roadblock polygons within 80\,m in front of the ego vehicle. The resulting region includes all lanes in the ego vehicle's direction of travel and all lawful branches it can take at intersections, while excluding sidewalks and lanes for oncoming traffic.

We then determine which image pixels provide reliable evidence about the ground. Using the camera calibration, each pixel ray is intersected with the estimated ground plane and retained only if the intersection is within 80\,m. Depth Anything~3 \citep{depthanything3} is then used to determine whether the ray is actually observing the ground. If the predicted depth is less than $70\%$ of the ground-intersection distance, the pixel is treated as a non-ground observation; otherwise, it is retained as ground evidence. The ground-plane height is estimated independently for each frame from the predicted depth. Ground behind annotated vehicles is retained so that the drivable-area label describes the underlying road surface rather than the temporary occupancy of that surface.

The drivable-area target $Y^{\mathrm{da}}_{k,c,n}$ is the fraction of the retained ground pixels in token $n$ that project inside the reachable road region. It is $1$ when all retained ground evidence lies within the reachable road, $0$ for verified non-drivable ground, and a soft value between $0$ and $1$ for tokens crossing a road boundary. Tokens for which fewer than $15\%$ of the pixels provide ground evidence, typically corresponding to sky, buildings, or other non-ground regions, are treated as unknown: their training target is set to $0$, and their contribution to the loss is reduced. Tokens receive full loss weight when ground evidence covers at least $30\%$ of their pixels. Below this threshold, the weight decreases linearly with the ground-evidence fraction but is clipped to a minimum of $0.1$. The weight is further multiplied by $3$ for boundary tokens with $0<Y^{\mathrm{da}}_{k,c,n}<1$. Frames for which no reachable road polygon can be constructed receive zero weight.

\paragraph{Collision-critical regions.}
The collision-critical label identifies objects that could cause a collision under a plausible ego motion, rather than marking every object in the scene. We begin with the predefined vocabulary of 16,384 ego trajectories from Hydra-MDP \citep{li2024hydra}, each spanning 4\,s at 10\,Hz. A trajectory is considered geometrically plausible only if all 40 poses remain inside the route-lane corridor expanded by 0.25\,m and its heading stays within $60^{\circ}$ of the nearest route-lane direction. We then retain trajectories with NAVSIM Driving Direction Compliance (DDC) equal to $1$. In the rare case that geometrically plausible trajectories exist but none satisfies this DDC condition, we use the geometrically plausible set as a fallback; scenes with no geometrically plausible trajectory simply have no positive collision-critical label.

Each retained trajectory is evaluated using the NAVSIM v1 simulator, which applies an LQR controller and a kinematic bicycle model to the ego vehicle while other agents replay their logged motion \citep{dauner2024navsim}. We define an object as collision-relevant if at least one retained trajectory incurs an at-fault collision with it under the NAVSIM v1 rules: a frontal collision, a collision with a stationary object, or a lateral collision occurring while the ego vehicle is outside its lane. The set of collision-relevant objects is determined once from rollouts initialized at $t_0$ and then tracked throughout the prediction horizon.

At each timestamp $t_k$, we retrieve the annotated 3D bounding box of every collision-relevant object present in that frame, clip the box against the camera near plane, and project it into each camera view. A token receives $Y^{\mathrm{col}}_{k,c,n}=1$ if its image patch overlaps at least one projected bounding box of a collision-relevant object, and $0$ otherwise. If a collision-relevant object is not annotated in a particular frame, it contributes no positive tokens at that timestamp. If the rollouts contain at-fault collisions but none of the corresponding collision-relevant objects is visible in any camera at $t_0$, the current-frame label is treated as unknown. Future timestamps at which no collision-relevant object is visible are likewise excluded from the loss and AP evaluation.

\section{Training Details}
\label{app:training}

\paragraph{Supervised fine-tuning.}
We train for 14{,}140 steps (35 epochs) with a per-device batch size of 2 on 128 NVIDIA H100 GPUs, giving a global batch size of 256. We use AdamW with gradient clipping at $1.0$, DeepSpeed ZeRO-2, and bf16 mixed precision. A single peak learning rate of $5\times10^{-5}$ is shared by all trainable parameter groups, with 404 warmup steps and cosine decay to $2.5\times10^{-7}$. The video VAE and the CLIP image encoder of Wan2.1-Fun-1.3B-InP \citep{wan2025}, as well as the vision encoder of the VLM backbone, remain frozen. 

\paragraph{Reinforcement learning setup.}
After SFT we optimize only the action DiT with GRPO \citep{shao2024deepseekmath} and keep the VLM, the world model, the learnable tokens, and all auxiliary heads frozen.

We construct the RL reward following the official NAVSIM scoring metrics, using the NAVSIM simulator to score each sampled trajectory on \texttt{navtrain}. For NAVSIM v1, we directly use PDMS as the reward. For NAVSIM v2, the Extended Comfort (EC) term in EPDMS depends on two consecutive frames and cannot be evaluated for an isolated trajectory. We therefore define 
\begin{equation}
r=\mathrm{NC}\cdot\mathrm{DAC}\cdot\mathrm{DDC}\cdot\mathrm{TLC}\cdot
\frac{5\,\mathrm{EP}+5\,\mathrm{TTC}+2\,\mathrm{LK}+2\,\mathrm{HC}}{14}
-\lambda_{\mathrm{ec}}\,\mathbf{1}\!\left[\text{EC violated}\right],
\end{equation}
with $\lambda_{\mathrm{ec}}=0.2$, where the EC flag is evaluated against the deterministic trajectory that the current policy produces for the preceding frame.

At each step, we sample 128 scenes and $G=16$ trajectories per scene, yielding 2{,}048 trajectories. We use a drift-corrected SDE flow-matching sampler with noise scale $\gamma_j=\eta\sqrt{1-u_j}$, which preserves the deterministic sampler's marginals. We adapt $\eta$ online (initialized at $0.3$ and clipped to $[0.1,0.6]$) to maintain useful reward variation. Advantages are mean-centered per scene and clipped to $\pm5$ without standard-deviation normalization; groups with a reward range below $0.02$ are skipped. We optimize the clipped policy-gradient loss ($\epsilon_{\mathrm{clip}}=0.2$) with a KL penalty to the frozen SFT policy ($\beta=0.01$). We set gradient clipping at $1.0$, and a learning rate of $3\times10^{-6}$ after 50 warmup steps. For each reward, we train for 4{,}500 steps on 8 NVIDIA H100 GPUs.

\section{More Visualization Results}
\label{app:more_vis}

\begin{figure}[t]
    \centering
    \includegraphics[width=\linewidth]{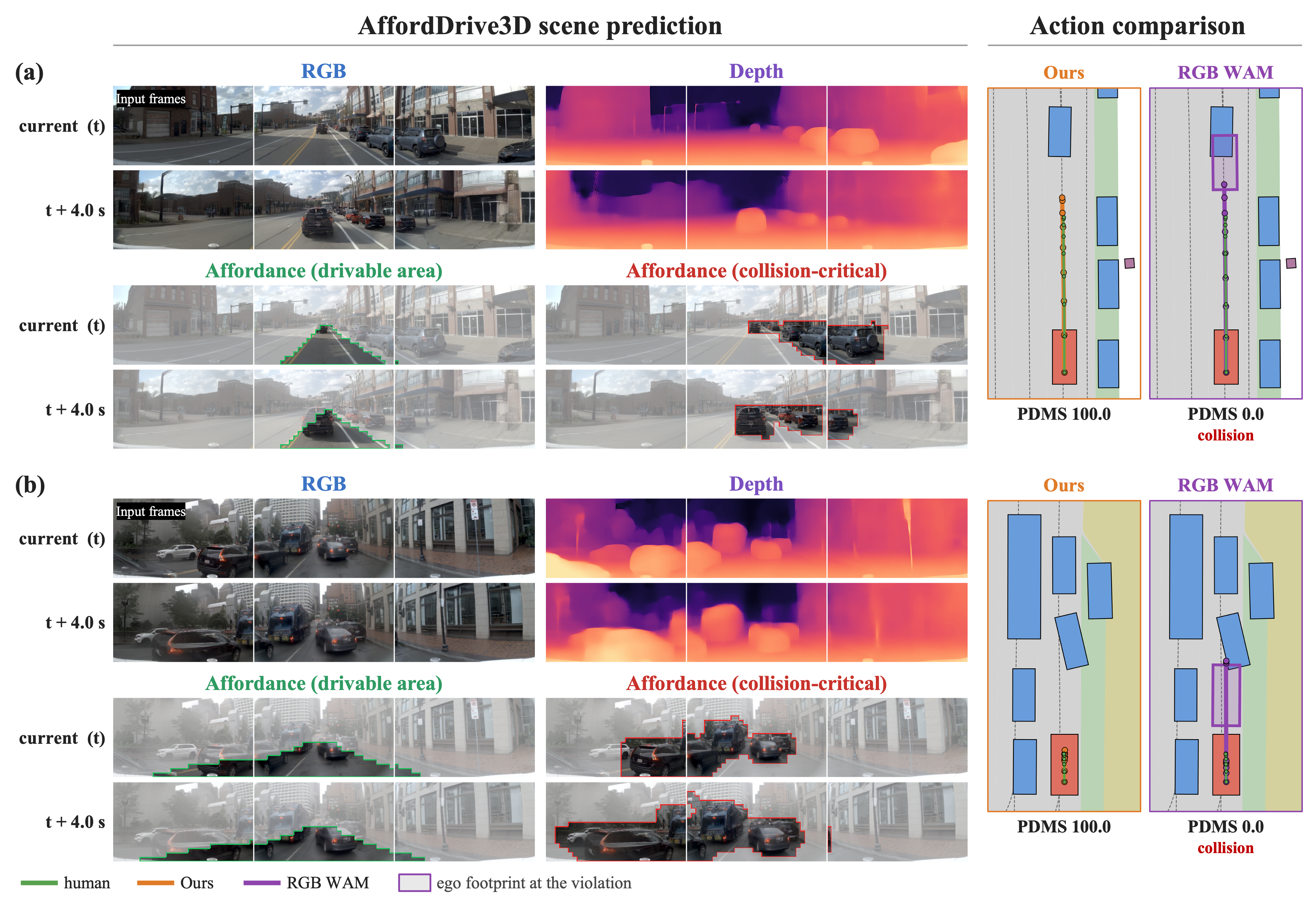}
    \caption{\textbf{Additional world-modeling and planning visualizations.} For each scenario, we show AffordDrive3D's predicted future RGB, current and future depth, and current and future drivable-area and collision-critical masks overlaid on the corresponding RGB views. The left, front, and right camera views are spatially concatenated, and the right panels compare the planned trajectories of AffordDrive3D and the RGB-based DriveVLA-W0 \citep{li2026drivevla}.}
    \label{fig:appendix-full-vis}
\end{figure}

\begin{figure}[t]
    \centering
    \includegraphics[width=\linewidth]{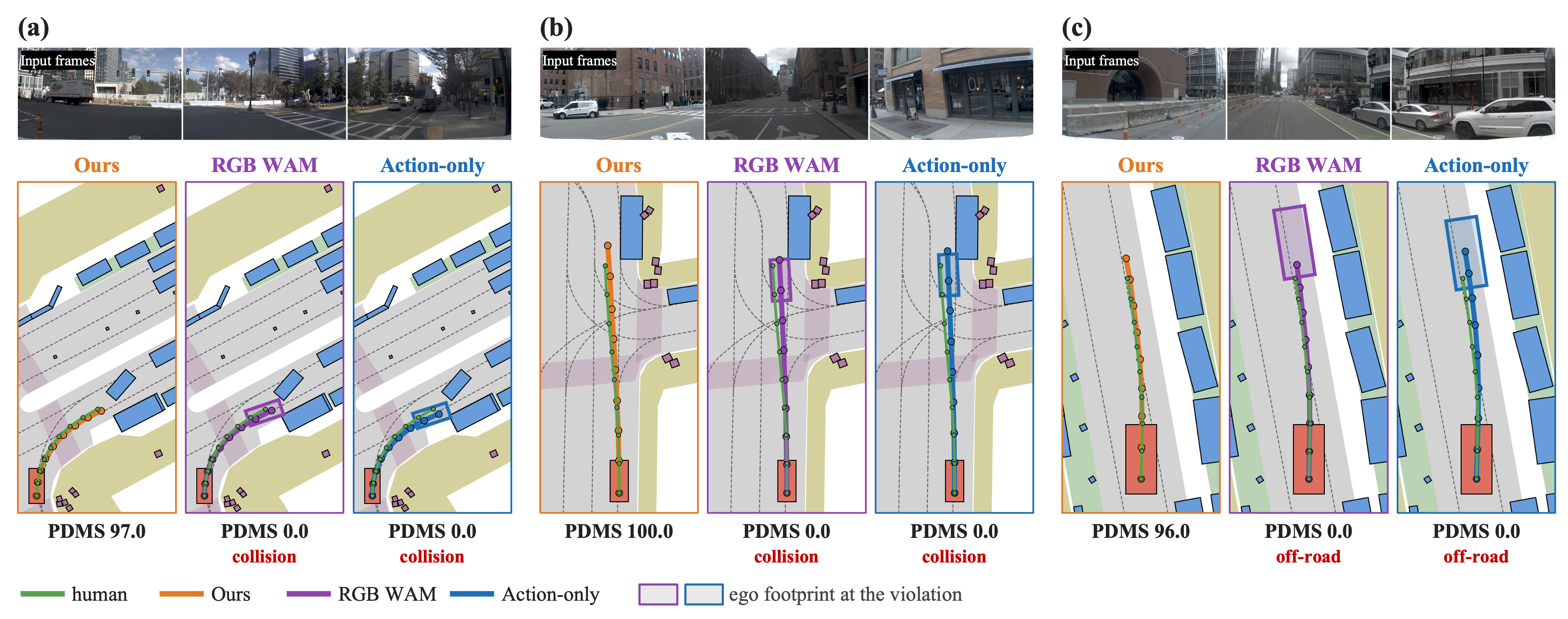}
    \caption{\textbf{Additional planning comparisons.} Each example shows the three input camera views and the BEV trajectories predicted by AffordDrive3D, the RGB-based DriveVLA-W0 \citep{li2026drivevla}, and the action-only VLA baseline.}
    \label{fig:appendix-traj-3cam}
\end{figure}

We provide additional qualitative results in \cref{fig:appendix-full-vis,fig:appendix-traj-3cam}. \Cref{fig:appendix-full-vis} visualizes the world-modeling outputs of AffordDrive3D, including predicted future RGB, current and future depth, and drivable-area and collision-critical masks overlaid on the corresponding RGB views. Following DriveDreamer-Policy \citep{zhou2026drivedreamer}, the left, front, and right camera views are spatially concatenated and modeled jointly, allowing scene content to transition consistently across views over time. \Cref{fig:appendix-traj-3cam} provides additional planning comparisons among AffordDrive3D, the RGB-based DriveVLA-W0 \citep{li2026drivevla}, and the action-only VLA baseline. For all BEV comparisons, the eight predicted ego poses are converted into simulated states at 10\,Hz using the NAVSIM LQR tracker and kinematic bicycle model. The highlighted ego footprint is drawn at the violation state, whereas surrounding agents are rendered from their $t=0$ annotations only for visual context.

\section{Affordance AP Computation}
\label{app:ap}

AP is computed independently for the collision-critical and drivable-area heads at each timestamp $t_k$ on \texttt{navtest}, where $t_0$ denotes the current frame. For each head and timestamp, we pool the predicted probabilities of all visual tokens across scenes and the $C=3$ cameras and compute standard non-interpolated AP directly from the ranked probabilities, without applying a prediction threshold.

For collision-critical prediction, scene--timestamp pairs with unknown labels are excluded, and all tokens from the remaining pairs use their binary targets. For drivable-area prediction, all tokens are included, and the soft target is binarized as positive when $Y^{\mathrm{da}}>0.5$. Tokens with less than $15\%$ ground evidence remain negative during evaluation, consistent with their training targets. 
\end{document}